\documentclass[10pt,twocolumn,letterpaper]{article}

\usepackage{times}
\usepackage{epsfig}
\usepackage{graphicx}
\usepackage{amsmath}
\usepackage{amssymb}
\usepackage{url}
\usepackage[pagebackref=false,breaklinks=true,letterpaper=true,colorlinks,bookmarks=false,breaklinks]{hyperref}

\usepackage{graphicx, color, xspace}
\graphicspath{{./}}
\usepackage{caption}
\usepackage{subcaption}
\usepackage{booktabs}

\usepackage{algorithm}
\usepackage{adjustbox}
\usepackage{diagbox}
\usepackage[noend]{algpseudocode}
\usepackage[utf8]{inputenc}
\usepackage{listings}
\usepackage{booktabs}

\title{Visual Wake Words Dataset}
\author{Aakanksha Chowdhery, Pete Warden, Jonathon Shlens, \\
Andrew Howard, Rocky Rhodes \\
Google Research \\
\{chowdhery, petewarden, shlens, howarda, rocky\}@google.com}
\date{}
\begin{document}
\maketitle
%\tableofcontents
%\newpage
%%%%%%%%% ABSTRACT
\begin{abstract}
The emergence of Internet of Things (IoT) applications
requires intelligence on the edge. Microcontrollers
 provide a low-cost compute platform to deploy intelligent IoT applications using machine learning at scale, but have extremely limited on-chip memory and compute capability.
 To deploy computer vision on such devices, we need tiny vision models that fit within a few hundred kilobytes of memory footprint in
terms of peak usage and model size on device storage. %Currently vision models are benchmarked on the CIFAR10 or ImageNet datasets both of which are restricted in terms of
%benchmarking the model accuracy and the memory costs for the common low-complexity microcontroller use-case.
To facilitate the development of microcontroller friendly models, we present a new dataset, Visual Wake Words, that represents a common microcontroller vision use-case of identifying whether a person is present in the image or not,
%The proposed dataset is derived from the publicly available COCO dataset, but
and provides a realistic benchmark for tiny vision models.
%that can be deployed on microcontrollers as edge devices with limited memory and compute capabilities.
Within a limited memory footprint of 250~KB, several state-of-the-art mobile models
achieve accuracy of 85-90\% on the Visual Wake Words dataset. We anticipate the proposed dataset will advance the research on tiny vision models that can push the pareto-optimal boundary in terms of accuracy versus memory usage for microcontroller applications.
\end{abstract}

\section{Introduction}
\label{sec:intro}
Deep learning has increased the accuracy of computer vision dramatically,
enabling their widespread use. Devices
at the edge have limited compute, memory, and power constraints; thereby requiring low-latency neural network architectures.
Recent advances have enabled real-time inference for several vision tasks on mobile platforms. First, CNN architectures, such as MobileNets~\cite{AHoward17, sandler2018mobilenetv2, howard2019mobilenetv3},
use operations such as depthwise separable
convolutions to optimize the latency with minimal
accuracy cost. Second, model compression techniques, such as
quantization~\cite{BJacob17} and pruning~\cite{zhu2017prune, IandolaMAHDK16} optimize the latency and size of the model
further. With the emergence of Internet of Things (IoT), the next frontier to deploy machine learning will be IoT sensors that use microcontrollers as limited compute platforms.

Microcontrollers are widely used for low-cost computation in IoT
devices ranging from industrial IoT to smart-home applications. Low-cost imaging sensors with microcontrollers
are economical to deploy in several vision applications,
for example, to sense if humans (or pets) are present in the building
or to monitor cars in garage or traffic cameras. The limited computation may not be sufficient
to get real-time insights of who or what is present.
However, if the object of interest is present, it can
then trigger an alert for human intervention or be streamed
to a network server or cloud for further analysis.

The two key challenges in deploying neural networks on microcontrollers
are the low memory footprint and the limited battery life.
In this paper, we focus on the low memory footprint. Typical microcontrollers have extremely limited
on-chip memory (100--320~KB SRAM) and flash storage (256~KB--1~MB).
The entire neural network model with its weight parameters and code has to fit within
the small memory budget of flash storage. Further,
the temporary memory buffer required to store the
input and output activations
during computation must not exceed the SRAM. Thus, we need to design tiny vision models that
achieve high accuracy on the typical microcontroller vision use-cases and fit within these constraints.

We set the following design constraints on the tiny vision
models for the microcontroller vision use-cases: the model size
and the peak memory usage must fit within a limited memory
footprint of 250~KB each; and the neural network computation must
incur less than 60~million multiply-adds per inference
at high accuracy. We observe that designing high-accuracy models under these constraints is challenging. Available image datasets are not representative of the
typical microcontroller vision use-cases. For example, the MobileNet V2 model architecture achieves only 58\% accuracy
at depth multiplier 0.5 with image resolution 160 on the ImageNet~\cite{deng2009imagenet} dataset
at 50~million multiply-adds and uses 1.95~million parameters.
However, ImageNet~\cite{deng2009imagenet} requires classification to a thousand classes and
this model could still be suitable for microcontroller vision use-case if it fit within the memory constraints. On the other hand, CIFAR10~\cite{krizhevsky2009learning} is a small dataset with a
very limited image resolution of 32-by-32 that does not accurately represent the benchmark model size or accuracy on larger image resolutions.

In this paper, we propose a new dataset, Visual Wake Words, as a benchmark for the tiny vision models to deploy on microcontrollers and advance research in this area.
We select the use-case of classifying images to two classes, whether a person is present in the image or not, as the representative vision task in the proposed dataset because it is one of the
popular microcontroller vision use-cases. This use-case is for a device to wake up when a person is present analogous to how audio wake words are used in speech recognition~\cite{warden2018speech}. This dataset filters the labels of the publicly available COCO dataset to provide a dataset of 115k training and validation images with labels of person and not-person.
We observe that this dataset provides valuable insights for model design because the accuracy on Visual Wake Words dataset can't be extrapolated directly from accuracy on the ImageNet dataset.
Further, the availability of full-resolution images in the proposed dataset allows users to train vision models at different image resolutions and benchmark the pareto-optimal boundary in terms of accuracy vs memory usage of the model design in this regime.

We discuss the memory-latency tradeoffs in deploying several
state-of-the art mobile models, such as MobileNet V1~\cite{AHoward17},
MobileNet V2~\cite{sandler2018mobilenetv2}, MNasNet (without squeeze-and-excite)~\cite{tan2018mnasnet}, ShuffleNet~\cite{zhang2018shufflenet} etc. in the tiny vision model regime.
In particular, the peak memory usage increases when the convolutional neural networks (CNNs) use residual blocks and inverted residual blocks. When the SRAM is limited, we process only a single path each time;
thus incurring additional latency when CNNs stage the computation of each path. Thus, the limited model size and peak memory usage of 250~KB determines the model parameters, such as selected image resolution, number of channels (depth multiplier for MobileNets), and the model depth. We conclude with a set of candidate models that achieve 85--90\% accuracy on the Visual Wake Words dataset and  define the pareto-optimal boundary in terms of accuracy vs memory usage.

\section{Related Work}

\begin{table}[tbp]
\small
    \centering
    \resizebox{\columnwidth}{!}{
        \begin{tabular}{l|c|c|c|c}
        \toprule[0.2em]
MCU Platform  &  Processor & Frequency  & SRAM & Flash  \\
   \toprule[0.2em]
FRDM-K64F~\cite{NXPmcureference} & Cortex-M4 & 120~MHz & 256~KB & 1~MB   \\ \hline
STM32 F723~\cite{STM32F7reference} & Cortex-M7 & 216~MHz & 256~KB & 512~KB \\ \hline
Nucleo F746ZG~\cite{STM32Nucleoreference} & Cortex-M7 & 216~MHz & 320~KB & 1~MB \\ \hline
NXP i.MX1060~\cite{NXPiMXprocessor} & Cortex-M7 & 600~MHz & 1~MB & External \\ \hline
 \end{tabular}
}
\caption{Some off-the-shelf microcontroller development platforms with their SRAM and flash storage constraints.
Note we list the microcontrollers with at least 250~KB SRAM, though other examples, such as, Cortex M0 only have 20--30~KB SRAM.}
\label{Table:MCUneeds}
\end{table}

\paragraph{Model compression.} Model architectures widely used and deployed on mobile devices include MobileNets~\cite{AHoward17}, and  MobileNetV2~\cite{sandler2018mobilenetv2}.  The proposed architectures have been further optimized for latency using hardware-based profiling~\cite{yang2018netadapt} and Neural architecture search~\cite{tan2018mnasnet, howard2019mobilenetv3}. ResNet-like model architectures~\cite{Resnetv2He, ResnetV1He} have also been optimized for mobile and embedded devices in ShuffleNet V2~\cite{zhang2018shufflenet, ma2018shufflenet} and SqueezeNet~\cite{iandola2016squeezenet}. Recent literature has also proposed several compression approaches that include quantization of weights and activations to leverage 8-bit arithmetic~\cite{BJacob17, krishnamoorthi2018quantizing}, and pruning the graph parameters that are less likely to affect accuracy \cite{zhu2017prune, IandolaMAHDK16, yang2018netadapt}.

\paragraph{Models for microcontroller devices.}
In the regime of microcontrollers, machine learning model design has focused on audio use cases such as keyword spotting. Recent papers benchmarking the speed-latency tradeoffs for the keyword-spotting use-case~\cite{warden2018speech}
on microcontrollers propose
20--64~KB model with 10-20 million operations per second~\cite{lai2018cmsis,lai2018rethinking,lai2018not,zhang2017hello}. Further speedups are potentially possible with the use of binary weights and activations, such as  XnorNet~\cite{rastegari2016xnor}, Trained Ternary Quantization~\cite{zhu2016trained} or with sparse architectures~\cite{fedorov2019sparse}. Under extreme memory constraints, recent work~\cite{tseng2018deterministic} has also proposed computing the convolution weight matrices on the fly using deterministic filter banks stored in the flash storage.

\section{Vision on Microcontrollers}
\subsection{Vision Use Cases}
\label{sec:vision-usecases}
Microcontrollers provide an economical way to deploy vision sensing at scale. Video
camera feeds, however, do not contain objects of interest at most times. For example, in building automation
scenarios, humans may be present in a small subset of image frames. Streaming the camera video feed
to the cloud continuously is not economical or useful. On-device sensing is economical when
the vision models deployed on the microcontrollers are tiny  and cost-effective.
%Microcontroller vision sensors are an effective low-cost option in IoT deployments
Microcontrollers are best suited for vision tasks where they act as `Visual WakeWords' to other sensors or generate alerts,
similar to the popular audio wakewords, such as ``Ok Google'' or ``Hey Google.''
We discuss some common use-cases for microcontroller vision models:

\paragraph{Person/Not-Person:} A popular use-case is sensing whether a person is present in the image or not.
Sensing person/not-person serves a wide-variety of use-cases including smart homes, retail, and
smart buildings. The inference result could then feed in to the If-this-then-that logic of various IoT
deployments enabling other IoT devices to start or trigger alerts. Generalizing from the person/not-person use case, low-cost vision sensors can
be economically deployed to sense the presence of specific objects of interest without high costs of inference,
%Some common use-cases for vision sensors include sensing the presence of pets
such as pets in a home, or cars in the garage.

\paragraph{Object counting:} In camera feeds, counting the number of objects provides a low-cost way to trigger other sensors. For example, in traffic monitoring, the vision sensors could count the number of cars and trigger alerts
in specific scenarios. %Similarly, in smart buildings or retail, vision sensors could count the number of persons, cars or objects on the shelf.

\paragraph{Object localization:}
Low-cost vision sensors can also be used to localize the object of interest, such as,
person, pet, car, or traffic signs and trigger useful alerts or reminders.
%For example, most cars today use a camera to sense if there is another car in the blind spot and trigger alerts.

\subsection{Microcontroller Systems}
\label{sec:design-constraints}
For the vision use cases in the previous section, we now
discuss the key challenges in deploying deep neural networks
to enable real-time inference on microcontrollers because
most microcontrollers are designed for embedded applications with low cost and do not have high throughput for
compute-intensive workloads such as CNNs.

\paragraph{On-chip memory footprint.} A typical microcontroller system consists of a processor core, an on-chip SRAM block and an on-chip embedded flash. One major constraint is the limited memory footprint in terms of on-chip memory and Flash storage.  Table~\ref{Table:MCUneeds} lists the details of some off-the-shelf microcontroller development
platforms with their SRAM and Flash storage.

The program binary, usually pre-loaded into the non-volatile flash, is loaded into the SRAM at
startup and the processor runs the program with the SRAM as the main data memory.
The size of the SRAM limits the size of memory that the software can use.
When the SRAM is extremely limited, we need to ensure that the peak memory usage of the model
computations is less than the total memory usage.

Peak memory usage is the maximum amount of total memory, including to store CNN activation maps, at any point in time during inference. Keeping activations in fast on-chip memory is important for performing fast inference, and large transient activations require large amounts of expensive local memory or long wait times if activations spill to slower off-chip memories. We assume that the weights will be read from the flash storage directly during the CNN inference, or else storing the temporary weights incurs additional memory usage.

As a first-order approximation, we estimate the peak memory usage as follows.
For each operation (e.g., matmul, convolution, pooling), we sum the size of the input allocations and output allocation. If the neural network is a simple chain of operations with no branching, for e.g. in MobileNet V1, select the maximum of these numbers and skip the following steps. For each parallel branch in the graph, for e.g. in MobileNet V2, we need to sum the activation storage of every pair of operations between the two branches. For each pair of operations, shared inputs must be counted once (e.g., the input to a simple residual block will be used in both parallel chains of the block) and we select the maximum of these numbers.

\paragraph{Model size.} To reduce the memory cost of storing the model on-device, the number of parameters in the model must be less than the flash memory storage. To save sufficient space for code of the compiler framework, we assume that 250~KB is needed to store the model in flash. Thus, the CNN model must be less than 250~K parameters assuming 8-bit representations for weights. %For example, Cortex-M4 and Cortex-M7 have integrated SIMD and MAC instructions that can be used to accelerate low-precision computation in neural networks.

\paragraph{Performance.} We benchmark the performance of CNNs
in terms of accuracy and multiply-adds per inference.
To enable at least one or more inferences per second on a microcontroller processor running at 100-200~MHz (see Table~\ref{Table:MCUneeds}), we limit the multiply-add operations on the microcontroller to be less than 60~million multiply-adds per inference. A CNN model with only 10~million multiply-adds per inference can increase the inference throughput by 5-10x.

The actual performance cost needs to be measured in terms of latency per inference because it includes the cost of memory accesses and the latency cost of convolutional kernels. Multiply-adds provide a first-order approximation of the inference cost agnostic to the
underlying platform architecture and kernel implementation specifics.

To ensure a long battery life for the IoT devices, we must also consider the energy efficiency of the microcontroller so that it can last without charging for long time periods. Energy efficiency depends on the computational cost of the CNN and the duty cycle
of processing; both of which continue to be an active area of research.

Based on the microcontroller system constraints in this section, we set the following design constraints on the tiny vision models for the microcontroller vision: the model size and the peak memory usage must fit within a limited memory  footprint of 250~KB each; and the CNN computation must incur less than 60~million multiply-adds per inference
at high accuracy.

%For each operation (e.g., matmul, convolution, pooling), sum the size of the input allocations and output allocation. If your neural network is a simple chain of operations with no branching, select the maximum of these numbers and skip the following steps.
%For each parallel branch in the graph, sum the activation storage of every pair of operations between the two branches. For each pair of operations, shared inputs can be counted once (e.g., the input to a simple residual block will be used in both parallel chains of the block). Select the maximum of these numbers.

%As a first order approximation, the FLOP count of the model
%in terms of multiply and add operations.  More accurate estimates requires designing models with
%more efficient ops based on the latency profiling of the device hardware~\cite{yang2018netadapt}.
 %Fig. XXX shows the
%latency profile of running some common operation using ARM CMSIS kernels on STM 32 F746NG.

%{\bf Energy efficiency.} The energy efficiency of the model depends on the computational cost of
%the operations in the model architecture and the frame-rate at which images are processed. Low-power
%model design will be an active area of research for microcontroller vision sensors.

\section{Vision Datasets for Microcontrollers}
\label{sec:datasets}
Currently vision models are benchmarked on the CIFAR10~\cite{krizhevsky2009learning} or ImageNet~\cite{deng2009imagenet} datasets both of which are restricted in terms of
benchmarking the model accuracy and the memory costs for the common low-complexity microcontroller use-case.
We present a new dataset, Visual Wake Words, that represents a common microcontroller vision use-case of identifying whether a person is present in the image or not,
The proposed dataset is derived from the publicly available COCO dataset, and provides a realistic benchmark for tiny vision models.
%We discuss the existing datasets used to benchmark vision models for microcontrollers
%and their shortcomings. Then, we discuss the details of our new proposed dataset,
%`Visual WakeWords Dataset,' for microcontrollers
%derived from COCO dataset and the steps required to build the new dataset.

\subsection{Existing Datasets}
\label{sec:existing-datasets}
%We discuss whether the existing datasets, ImageNet and CIFAR10, are suitable to benchmark the
%microcontroller vision use-cases.

ImageNet~\cite{deng2009imagenet} is the most widely used dataset for vision benchmarks. It constitutes
tens of millions of annotated images with labels that
%organized by the semantic hierarchy of WordNet with an average of 500-
%1000 full-resolution images for each synset. The image classification task
classify the images in to 1000 classes providing a generic benchmark for several state-of-the-art models.
There are several challenges in using this dataset to benchmark the microcontroller vision use-cases. First, ImageNet dataset does not address a very popular use-case of person/not-person classification because it does not
have instances of person class. Second, the typical vision use-cases for microcontroller do not require a large number
of classes. The parameters of the last few layers of CNN in the classification model can dominate the model size. For a binary classification task, the model size can be compressed
substantially by reducing the memory footprint of last few layers.

CIFAR10 dataset~\cite{krizhevsky2009learning} has been actively used to benchmark some recent  microcontroller vision models. For example, recent work~\cite{zhang2017hello} benchmarks a 5-layer convolutional neural network
(CNN) for CIFAR-10 dataset on an off-the-shelf Arm Cortex-M7 platform classifying 10.1 images per second with an accuracy of 79.9\%. However, CIFAR10 dataset is a small dataset that consists of 60000 32x32 color images in 10 classes. Limited image resolution limits its use to benchmark model accuracy for larger image resolutions. Further the model size and memory usage
scales with the square of the image resolution and thus, the accuracy benchmarks on this dataset are not representative of the peak memory usage or the model size if the model will be used for larger image resolutions.

\subsection{Visual Wake Words Dataset}
\label{sec:new-dataset}
We define a new dataset to fit a popular use-case for microcontrollers discussed in
Section \ref{sec:vision-usecases}. We focus on the binary classification use-case where the images are labeled
with two labels: object-of-interest is present or not present. The most interesting use-case enables the microcontrollers
to classify whether there is a person in the image or not. Hence, we call this the Visual Wake Words dataset because it enables
%it allows the image frames with a person to act as a visual wake word to generate alerts or trigger other processing tasks similar to the scenario is for
a device to wake up when a human is present analogous to how audio wake words are used in speech recognition~\cite{warden2018speech}.

We derive the new dataset by re-labeling the images available in the publicly available COCO dataset with labels
corresponding to whether the object-of-interest is present or not. This provides a powerful way to benchmark this common
vision use-case for microcontrollers with a simple tweak to the COCO dataset.
COCO dataset~\cite{lin2014microsoft} is widely used to benchmark object detection and segmentation tasks.
COCO dataset comprises natural images of complex everyday scenes
that contain multiple objects and it has 91 objects types with more than a million labeled instances in 115k images in the training and validation set.

The process of creating new labels for Visual Wake Words dataset from COCO dataset is as follows.
Each image is assigned a label 1 or 0. The label 1 is assigned as long as it has at least one bounding box
corresponding to the object of interest (e.g. person)
with the box area greater than 0.5\% of the image area.
In the person/not-person use-case, the label 1 corresponds to the `person' being present in the image and the label 0
corresponds to the image not containing any objects from
person class. To generate the new annotations and the files in TensorFlow records format corresponding to this dataset, use the script {\tt  \footnotesize build\_visualwakewords\_data.py} available in open-source in {\tt  \footnotesize `tensorflow/models/research/slim/datasets'}~\cite{VisualWakeWordsDataset2019}.
%~\ref{SlimModel17}
%\ref{VisualWakeWordsDataset2019}

The command to generate the Visual Wake Words dataset from {\tt  \footnotesize `tensorflow/models/research/slim'} directory using the script %'\verbatim{build_visualwakewords_data.py}' is
as follows:
\begin{lstlisting}[language=Python]
python datasets/build_visualwakewords_data.py \
 --train_image_dir="${TRAIN_IMAGE_DIR}" \
 --val_image_dir="${VAL_IMAGE_DIR}" \
 --test_image_dir="${TEST_IMAGE_DIR}" \
 --train_annotations_file="${TRAIN_ANNOTATIONS}" \
 --val_annotations_file="${VAL_ANNOTATIONS}" \
 --output_dir="${OUTPUT_DIR}" \
 --small_object_area_threshold=0.005
\end{lstlisting}

\begin{figure}
    \centering
    \begin{subfigure}[b]{0.45\columnwidth}
        \includegraphics[width=\textwidth]{COCO_val2014_000000000338.jpg}
        \caption{`Person'}
        \label{fig:person}
    \end{subfigure}
    ~
    \begin{subfigure}[b]{0.45\columnwidth}
        \includegraphics[width=\textwidth]{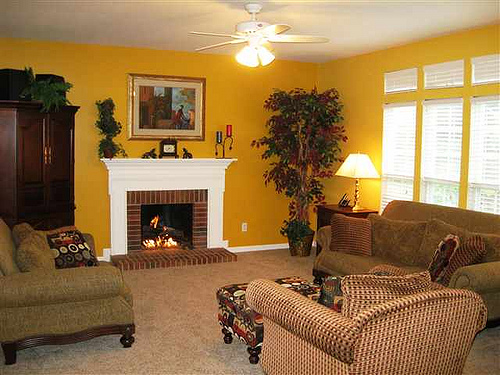}
        \caption{`Not-person'}
        \label{fig:notperson}
    \end{subfigure}
    \caption{Sample images labeled to `person' and `not-person' categories from COCO training dataset.}
    \label{fig:person-not-person}
\end{figure}

Figure~\ref{fig:person-not-person} illustrates an example of
sample images labeled to `person' and `not-person' categories from the COCO training dataset.

The command can be modified to classify the presence of another object of interest instead of the `person'
class by adding an additional flag specifying the new label from the COCO category. The default is set to the `person` category.
\begin{lstlisting}[language=Python]
 --foreground_class_of_interest=`person'
\end{lstlisting}

We use the above script on the training and validation image set of COCO2014 dataset~\cite{lin2014microsoft}. We use a total of 115k images of the training and validation dataset to generate new annotations.
We assign `person' label when the image contains at least one bounding box from person class with the bounding box area greater than $0.5$\% of the image area. We observe that the reassigned labels are roughly balanced between the two classes: 47\% of the images in the training dataset
of 115k images are labeled to the `person' category, and similarly, 47\% of the images in the validation dataset are labeled  to the `person' category. For evaluation, we use the minival~\cite{COCO-minival, huang2017speed} on the COCO2014 dataset~\cite{lin2014microsoft} comprising a total of 8k images.

\begin{figure*}[tbhp]
    \centering
  \includegraphics[width=\linewidth]{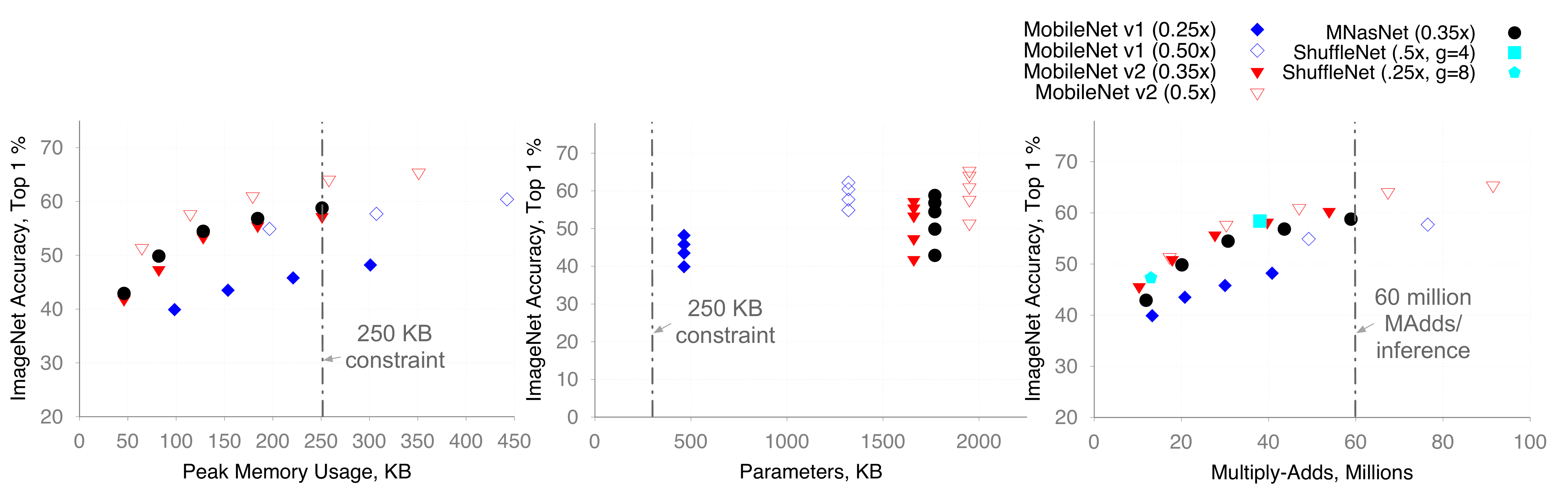}
\caption{On ImageNet dataset, we compare the top-1 accuracy versus the memory footprint, model size and multiply-adds per inference. Figure (a) shows the top-1 accuracy vs estimated peak memory usage (in KB), (b) the top-1 accuracy vs number of parameters (in KB), and figure (c) shows the top-1 accuracy vs multiply-adds (in millions). Each point corresponds to different image resolution in \{96, 128, 160, 192, 224\}.}
\label{fig:imagenet-accuracy}
\end{figure*}

\section{Experiments}
We evaluate the tiny vision models
on the ImageNet dataset and the proposed Visual Wake Words Dataset for
the person/not-person classification task.
We present experimental results to demonstrate the effectiveness
of the proposed dataset to benchmark the microcontroller
vision use-cases.
We benchmark the following models: MobileNet V1~\cite{AHoward17}, MobileNet V2~\cite{sandler2018mobilenetv2}, MNasNet (without squeeze-and-excite)~\cite{tan2018mnasnet}, and Shufflenet~\cite{zhang2018shufflenet}.
We compare accuracy versus measures of resource usage such as the peak memory usage, model size (number of parameters) and inference multiply adds. Our design goal is tiny vision models that use less than 250~KB in SRAM, the model size fits within 250~KB of available flash storage, and the inference cost is less than 60~million multiply-adds per inference.

%The experiments evaluate the following key questions: a) should I lower the image resolution or the depth multiplier to fit the vision model in limited memory, b) how does the accuracy when the model is trained on ImageNet dataset vs the Person/Not-Person dataset, and c) how does the latency change with different implementations of convolution layers on the Cortex M7 microcontroller.

%\subsection{Person/Not-Person Classification}
\subsection{Training setup}
MobileNet and MNasNet models are trained in TensorFlow using asynchronous training on GPU using the standard RMSPropOptimizer with both decay and momentum set to 0.9. We use 16 GPU asynchronous workers, and a batch size of 96. We use an initial learning rate of 0.045, and learning rate decay rate of 0.98 per epoch.  All the convolutional layers use batch normalization with average decay of 0.99. Using the floating-point checkpoints, we then train models with 8-bit representation of weights and activations by using quantization-aware training~\cite{TFQuantize, BJacob17} with a learning rate of $10^{-5}$ and learning rate decay of 0.9 per epoch. Since we only classify two classes in Visual Wake Words dataset,
we shrink the last convolutional layer in MobileNet V2 and MNasNet models.

In the Visual Wake Words dataset, we evaluate the accuracy on the minival image ids~\cite{COCO-minival,huang2017speed} of COCO2014 dataset~\cite{lin2014microsoft} comprising a total of 8k images, and for training, we use the remaining images out of 115k images in training/validation dataset (not including the 8k COCO2014 minival images).

\begin{figure*}[tbhp]
    \centering
   \includegraphics[width=\linewidth]{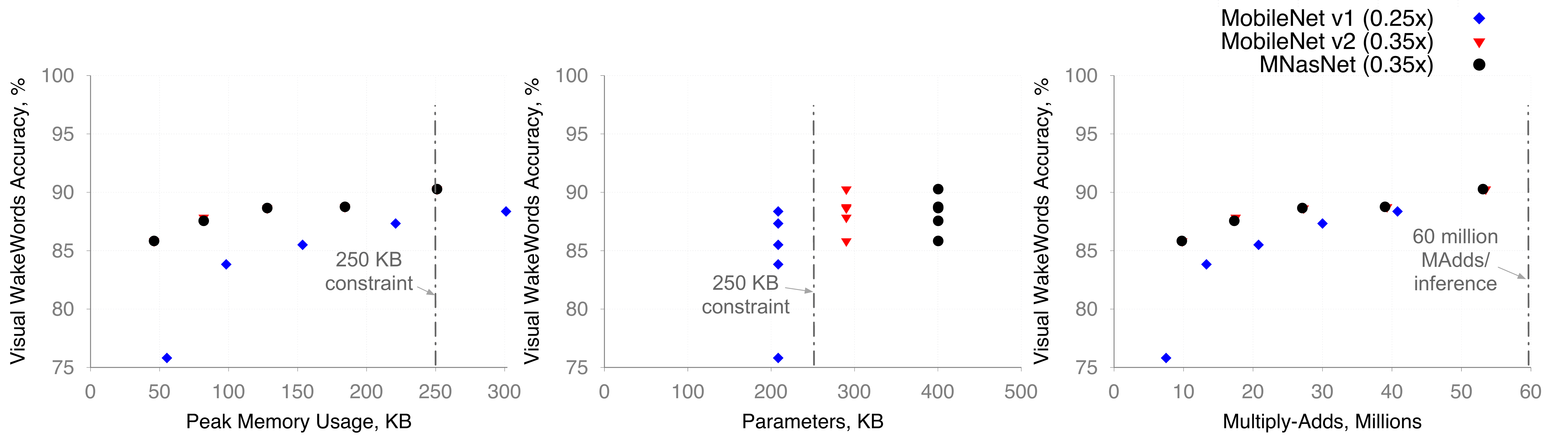}
\caption{On Visual Wake Words dataset, we compare the accuracy versus the memory footprint, model size and multiply-adds per inference. Figure (a) shows the accuracy vs estimated peak memory usage (in KB), (b) the accuracy vs number of parameters (in KB), and figure (c) shows the accuracy vs multiply-adds (in millions). Each point corresponds to different image resolution in \{96, 128, 160, 192, 224\}. Note that the red and black points are overlapping.
}
\label{fig:visualwakewords-accuracy}
\end{figure*}

\subsection{Accuracy}
We compare and contrast the accuracy of these models on ImageNet dataset vs
the Visual Wake Words Dataset.
Figure~\ref{fig:imagenet-accuracy} illustrates the top-1 accuracy of MobileNet~\cite{AHoward17}, MNasNet (without squeeze-and-excite)~\cite{tan2018mnasnet}, and ShuffleNet~\cite{ma2018shufflenet} models on ImageNet dataset (that classifies the images to 1000 classes) with respect to peak memory usage, model size, and multiply-adds. We measure the model size in terms of number of parameters in kilobytes assuming 8-bit representation of weights. We vary the depth multiplier of the models and for each depth multiplier, we illustrate the model accuracy at multiple image resolutions in \{96, 128, 160, 192, 224\}. With fixed inference cost of 50~million multiply-adds, the accuracy improves in the order of MobileNet V1, MobileNet V2, ShuffleNet, and MNasNet (without squeeze-and-excite). Note that the top-1 accuracy increases with increased depth multiplier and image resolution.
For a fixed model size, we can reduce the peak memory usage and the multiply-adds by reducing the image resolution. For example, increasing the image resolution from 96 to 224 increases the top-1 accuracy of MobileNet V1 with depth multiplier 0.25 from 39.9\% to 48\%, and of MNasNet from 42.91\% to 58.79\%.
However, there is a trade-off in terms of the peak memory usage, number of parameters, and multiply-adds.
For example, in the regime of models with less than 60~million multiply-adds, MobileNet V1 model with depth multiplier 0.5 on image resolution 128 achieves higher top-1 accuracy of 54.9\% compared to the same model with depth multiplier 0.25 on image resolution 224. However, it is not feasible to fit this model in 250KB because the model trained on ImageNet dataset has much higher model size and peak memory usage.

Figure~\ref{fig:visualwakewords-accuracy} illustrates the accuracy of correct classification on the
two classes person/not-person on the Visual Wake Words dataset. Note that the top-1 accuracy of
MobileNet V1, MobileNet V2 and MNasNet (without squeeze-and-excite) is less than 60\% on the ImageNet dataset, however, the accuracy is greater than 85\% for the three models on the Visual Wake Words dataset.
Further, we note that all models
fit within the peak memory usage of 250~KB. However, MobileNet V2 and MNasNet models require
special memory management to fit the model within these constraints that we discuss in detail in the next section.
In terms of model size (number of parameters) in KB, MobileNet V1 with depth multiplier 0.25
is 208 KB while the MobileNet V2 and MNasNet models with depth multiplier 0.35 require 290~KB and
400~KB respectively. This suggests that model size can be compressed further by
reducing the depth multiplier and additional techniques such as pruning to provide smaller
model sizes.

%\begin{figure}[tbhp]
%    \centering
%        \includegraphics[width=.95\columnwidth]{figures/accuracy_vs_madds.eps}
%        \caption{ImageNet dataset: Top-1 Accuracy vs Multiply-adds (Millions).}
%        \label{fig:latencyprofile}
%\end{figure}
%\begin{figure}[tbhp]
%    \centering
%        \includegraphics[width=.95\columnwidth]{figures/accuracy_vs_peak_memory.eps}
%        \caption{ImageNet dataset: Top-1 Accuracy vs Estimated Peak Memory Usage (in KB).}
%        \label{fig:memoryusage}
%\end{figure}
%\begin{figure}[tbhp]
%    \centering
%        \includegraphics[width=.95\columnwidth]{figures/accuracy_vs_peak_memory_VWW.eps}
%        \caption{Visual Wake Words Dataset: Accuracy vs Estimated Peak Memory Usage (in KB).}
%        \label{fig:memoryusage}
%\end{figure}

\subsection{Memory-latency trade-offs}
\paragraph{Peak Memory Usage.}
We use on-chip SRAM to store the input/output activation maps of each layer assuming that we reuse the temporary buffer space for storing the input and the output buffer space. To benchmark the peak memory usage, we assume 8-bit representation for activations.
We find that the peak memory usage is often dominated by the input and output activation maps of the first few layers in MobileNet and MNasNet architectures.
To limit the peak memory usage, the output channels in first convolutional layer are limited to eight channels when we use depth multiplier 0.35 for MobileNet V2 and MNasNet model architectures.

\begin{figure}
    \centering
        \includegraphics[width=.9\columnwidth]{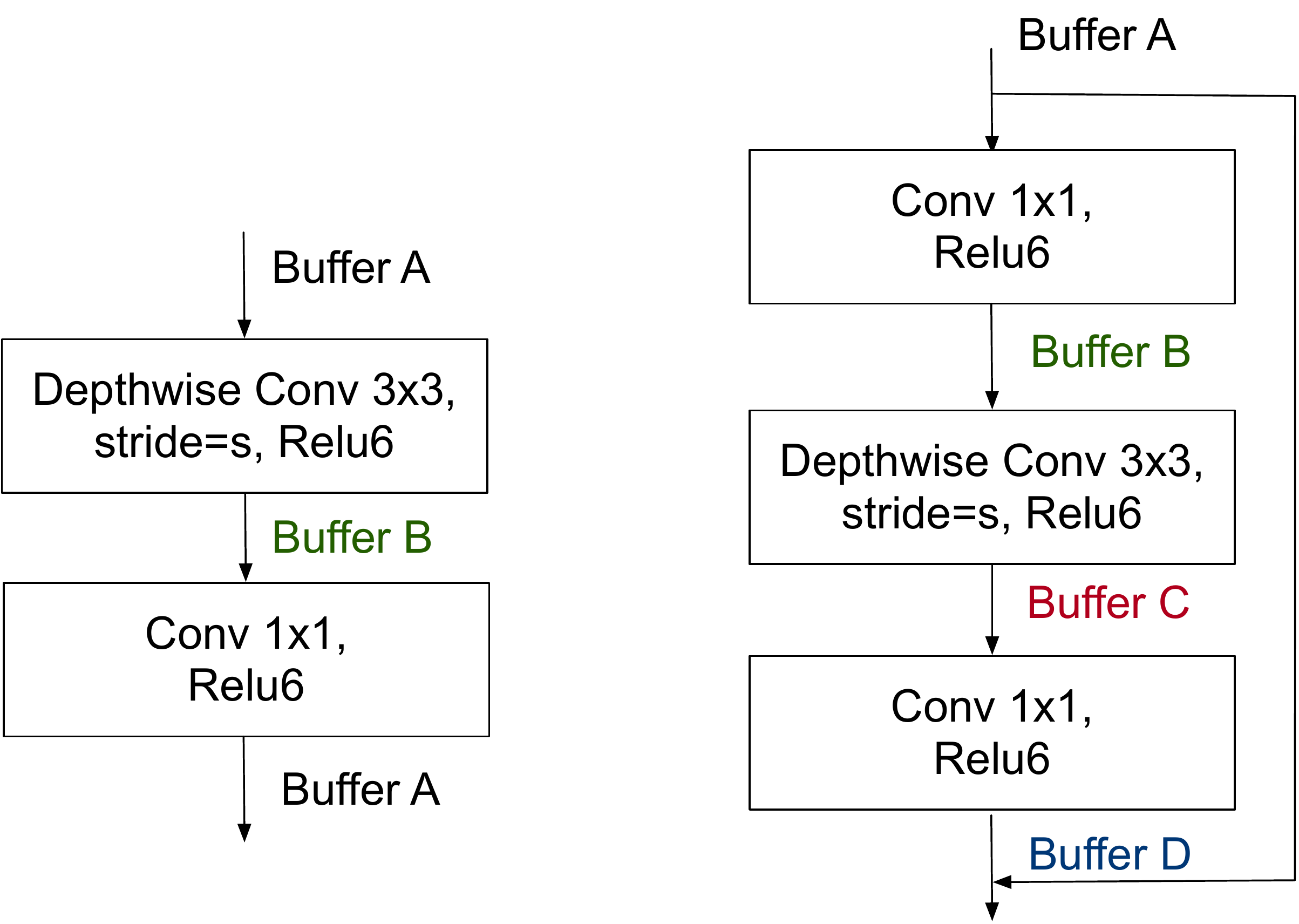}
        \caption{Temporary buffer management for MobileNet V1 (left) and MobileNet V2 (right).}
        \label{fig:bufferprofile}
\end{figure}

Figure~\ref{fig:bufferprofile} illustrates this for two different models: MobileNet V1 and MobileNet V2.
Note that in MobileNet V1 architecture, we re-use the buffer $A$ to store the inputs to conv 1x1 layer.
In MobileNet V2 architecture, we have parallel paths between the inputs and the outputs of different layers,
but the projection and expansion layers do not need to be materialized completely using tricks suggested in~\cite{sandler2018mobilenetv2}. We assume that the expansion/depthwise-convolution/projection matrices are split into six parallel paths and each path is materialized only during its computation. As a result, we need additional temporary buffers $B$ and $C$ to store the intermediate values so that they can be added to the output afterward.

Peak memory usage for each of the neural-network layers must fit within the SRAM. We observe in Figure~\ref{fig:visualwakewords-accuracy} that for a given model architecture with a fixed depth multiplier, scaling the input image resolution reduces the peak memory usage.
%, where we set the target of less than 250 KB in size.
In the first few layers, the activation map size is directly proportional to
the square of image resolution that will be a limiting factor in selecting the architecture and its number of channels.
In the last few layers, the larger number of channels could dominate the memory required. Further note that the convolution kernel may
require additional buffer space to store the temporary outputs when using GEMM kernels that affects the memory consumption of the last few layers. Further optimization requires efficient convolution kernels that are computed in-place or in a tiled implementation, further reducing the memory usage of
specific layers.

\paragraph{Model size.} The model must fit in the flash memory typically within a few hundred kilobytes (up to 1~MB). Techniques to reduce the model size include using a lower depth multiplier in the MobileNet type architectures, and compression techniques, such as quantization or pruning.

Typically models with a higher depth multiplier achieve higher accuracy but require a larger model size. Reducing the input image resolution will only reduce the peak memory usage or latency but not the model size. Consider the scenario with image resolution of 224, MobileNet V2 architecture with depth multiplier 0.35 requires 290KB and with depth multiplier 0.5 requires 569KB to store weights of the layers that extract features in 8-bit.

\paragraph{Latency.} Model performance depends on the computational cost of the convolution kernels.
%Assume that we deploy model with size constraints of 250KB and would
%like to process images at 0.5-1 frames/sec. Then, a first-order approximation is limit the multiply-and-add
%operations of these kernels between 10 and 50 million operations.
MobileNet architectures' latency is dominated by pointwise and separable depthwise convolution kernels, that can be made faster using
fixed-point arithmetric with 8-bit representation of models or using GEMM kernels. In some cases, we may choose to trade-off memory for latency,
for example, we can compute depthwise separable convolutions in-place using for loops to save memory footprint, but then we can no longer use the memory rearrangement techniques to speed up the kernels. Alternate implementations of convolution kernels using lower-precision weights could speed up the convolution kernels.

%\paragraph{Resource consumption on Visual Wake Words.}
%In Visual wake words, we classify the final logits to two classes: person and not-person and therefore, the last convolutional layer of CNN to predict logits is very small. So, the number of parameters and the number of Multiply-add operations correspond to the feature extraction layers of the CNNs. To benchmark the peak memory usage, we assume 16-bit floats for activations.
%Assuming that the temporary buffer can be reused, we only need to store the input and output activation maps of each CNN layer. For MobileNet V1 and ShuffleNet, we need to store the entire input and output activation maps, but for MobileNet V2 and MNasNet, the memory requirement is reduced since the non-bottleneck tensors do not need to be materialized. We compare the number of parameters, number of Multiply-Adds and the peak memory usage in Table~\ref{Table:QtzAwareTrn_madd}. Thus, MobileNet V2 and MNasNet architectures work better when SRAM  memory is the main constraint.

%\paragraph{Latency Profile on ARM Cortex M7.}
We profile the latency of deploying the convolutional layers of the MobileNet V1 model with depth multiplier (0.25x) on the microcontroller
development board STM32 F746~\cite{STM32F7reference}. We use ARM
CMSIS5-NN kernels to program the depthwise and pointwise convolutions in MobileNet V1
architecture.
%Figure~\ref{fig:latencyprofile} illustrates the variation in latency across all the convolutional layers on image resolution of 224-by-224. Per-layer latency varies from 70-150ms. Note that
The overall latency per inference is approx 1.3~sec enabling approximately 0.75~frames/sec. This frame rate suffices in application requirements when the person count doesn't change frequently.

%\begin{figure}
%    \centering
%        \includegraphics[width=.95\columnwidth]{figures/latency_chart.png}
%        \caption{Latency of running convolutional kernels in MobileNet V1 (0.25x) on microcontroller STM 32 F7 development board.
%        We use ARM CMSIS5-NN kernels to benchmark the latency values.}
%        \label{fig:latencyprofile}
%\end{figure}

% For ShuffleNet, we use 2x, g =
% 3 that matches the performance of MobileNetV1 and
% MobileNetV2. For the first layer of MobileNetV2 and
% ShuffleNet we can employ the trick described in Section 5 to reduce memory requirement. Even though
% ShuffleNet employs bottlenecks elsewhere, the non bottleneck tensors still need to be materialized due to the
% presence of shortcuts between non-bottleneck tensors.

%\newpage
\section{Conclusions and Future Work}
\label{sec:conclusions}
In this paper, we introduced the challenges in deploying neural networks on
microcontrollers that can enable intelligence on the edge at scale in several
emerging IoT applications. Limited on-chip memory will be a major constraint
in deploying CNN models on microcontrollers. We believe that such model deployments
will require tiny vision models fitting both the model size
and the peak memory usage within 250~KB at less than 60~million multiply-adds per inference.
Existing datasets are not representative of the pareto-optimal boundary of model
accuracy vs peak memory usage for such tiny vision models.
To advance research in this area, we present a new dataset, `Visual Wake Words',
that classifies images for the presence of a person or not that is
a popular microcontroller vision use-case. Our initial benchmarks on the state-of-the-art
models with 8-bit weights and activations on this dataset achieve up to 90\% for classification task,
and the same MobileNet models can also form the basis of person counting and localization tasks.
We believe that the Visual Wake Words Dataset will provide researchers with a
platform to fundamentally rethink the design of tiny vision models that fit on microcontrollers.

%\newpage
\bibliographystyle{unsrt}
\bibliography{bibliography}

\end{document}